\documentclass[lettersize,journal]{IEEEtran}
\usepackage{amsmath,amsfonts}
\usepackage{algorithmic}
\usepackage{algorithm}
\usepackage{array}
\usepackage[caption=false,font=normalsize,labelfont=sf,textfont=sf]{subfig}
\usepackage{textcomp}
\usepackage{stfloats}
\usepackage{url}
\usepackage{verbatim}
\usepackage{graphicx}
\usepackage{cite}
\usepackage{graphicx}
\usepackage{multirow}
\usepackage{amsmath,amssymb,amsfonts}
\usepackage[thmmarks]{ntheorem}
\begin{document}
\bibliographystyle{IEEEtran}

\title{Reviewing continual learning from the perspective of human-level intelligence}

\author{Yifan Chang,
		Wenbo~Li,
        Jian Peng,
        Bo Tang,
        Yu Kang,
        Yinjie Lei,
        Yuanmiao Gui,
        Qing Zhu,
        Yu Liu,
        Haifeng Li~\IEEEmembership{Member,~IEEE,}
\thanks{This work was supported by the National Nature Science Foundation of China under Grant 41871302, the Development Program of China under Grant 2018YFB1004600 and Anhui Provincial Natural Science Foundation under Grant 2108085J19. Corresponding author: Haifeng Li, email: lihaifeng@csu.edu.cn.
}

\thanks{Y. Chang and Y. Kang are with  with the State Key Laboratory of Fire Science, Department of Automation and Institute of Advanced Technology, University of Science and Technology of China, Hefei, China.}
\thanks{W. Li and Y. Gui are with Institute of Intelligent Machines, Chinese Academy of Sciences, Hefei 230031.}
\thanks{B. Tang is with the the Department of Electrical and Computer Engineering at Mississippi State University, USA.}
\thanks{Y. Lei is with the college of Electronics and Information Engineering at Sichuan University,  Chengdu, China.}
\thanks{Q. Zhu is with the Faculty of Geosciences and Environmental Engineering of the Southwest Jiaotong University, Chengdu, China.}
\thanks{Y. Liu is with the Institute of Remote Sensing and Geographic Information System, Peking Universit, Beijing, China}
\thanks{J. Peng and H. Li.are with the School of Geosciences and Info-Physics, Central South University, South Lushan Road, Changsha, 410083, China.}

}

\markboth{Journal of \LaTeX\ Class Files,~Vol.~14, No.~8, August~2021}%
{Shell \MakeLowercase{\textit{et al.}}: A Sample Article Using IEEEtran.cls for IEEE Journals}


\maketitle

\begin{abstract}
Humans continual learning (CL) ability is closely related to Stability Versus Plasticity Dilemma that describes how humans achieve ongoing learning capacity and preservation for learned information. The notion of CL has always been present in artificial intelligence (AI) since its births. This paper proposes a comprehensive review on CL. Different from previous reviews that mainly focus on the catastrophic forgetting phenomenon in CL, this paper surveys CL from a more macroscopic perspective based on the Stability Versus Plasticity mechansim. Analogous to biological counterpart, "smart" AI agents are supposed to i) remember previously learned information (information retrospection); ii) infer on new information continuously (information prospection:); iii) transfer useful information (information transfer), to achieve high-level CL. According to the taxonomy, evaluation metrics, algorithms, applications as well as some open issues are then introduced. Our main contributions concern i) recheck CL from the level of artificial general intelligence; ii) provide a detailed and extensive overview on CL topics; iii) present some novel ideas on the potential development of CL.
\end{abstract}


\section{Introduction}
\IEEEPARstart{W}{hat} is intelligence? Many theories \cite{russell2002artificial} have been developed to define it, for example, intelligence is the resultant of the process of acquiring, storing in memory, retrieving, combining, comparing, and using in new contexts information \cite{humphreys1979construct}; intelligence is goal-directed adaptive behavior \cite{eisen1982handbook} and so on \cite{feuerstein2002dynamic}. While different in terms of perspective, a common and central idea can be noticed: the ability to mold our cognitive system to deal with the always changing demanding circumstances \cite{schulz2012origins}, which is strictly related to the ability of continual learning (CL).

\begin{figure}[!t]
\centering
\includegraphics[width=3in]{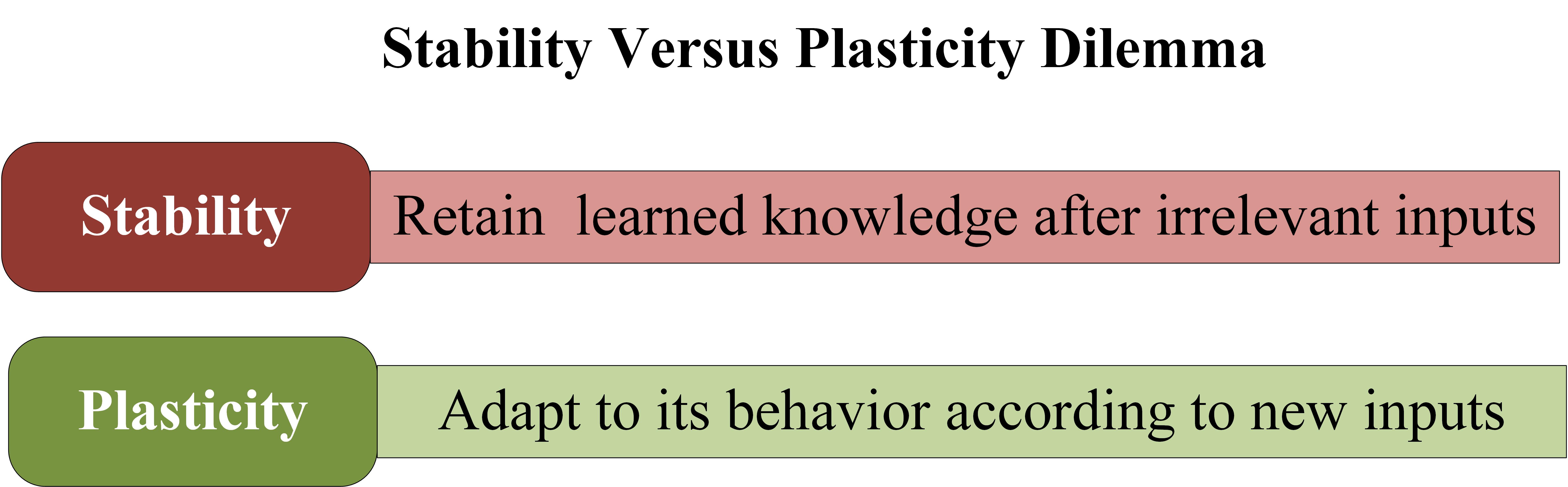}
\caption{The interpretion of Stability Versus Plasticity Dilemma [8].}
\label{img1}
\end{figure}

Humans have the extraordinary ability to learn continually from experience. Not only can we remember previously learned knowledge and apply these skills to new situations, we can also abstract useful knowledge and use these as the foundation for later learning. The CL ability is closely owed to Stability Versus Plasticity Dilemma which explores how a learning system remains adaptive in response to input, yet remains stable in response to irrelevant input \cite{bedia2009bio}. Stability is the ability to retain the learned information on the old tasks and plasticity is the ability to adapt to a new task \cite{abraham2005memory}, as shown in Figure \ref{img1}.

Analogous to our biological counterpart, achieving adaption is also a grand goal of artificial intelligence (AI) \cite{ring1994continual}. The aim is to build an artificial CL agent that adapts to a sophisticated understanding of the world from its own experience through the incremental development of skills and knowledge \cite{ring1994continual}.

Up to present, many progress has been witnessed in CL field and many paper have summarized CL, for example, Van et al.  \cite{van2019three}, Thrun et al. \cite{thrun1995lifelong} and Farquhar et al. \cite{farquhar2018towards} surveyed on certain aspects of CL, such as applications and evaluation metrics; Parisi et al. [18] provided a comprehensive review on the catastrophic forgetting problem. However, these reviews mainly focus on a specific issue on CL. In fact, to achieve the aim of artificial general intelligence (AGI), an ideal CL agent is supposed to perform well in the whole process of CL, not only on the forgetting problem for the past encountered information. To bridge this gap, this paper reviews CL on a macroscopic level based on the Stability Versus Plasticity mechansim. The advantages of examining CL in the context of human-level intelligence lie in two reasons: on the one hand, AI agents are conceptually derived from the biological neural networks, and thus, such an examination may facilitate the validation. What's more important, it can clearly show how far the current work is from the final destination in CL, providing introspection for existing research while serving as inspiration for future work.

\begin{figure*}
\centering
\includegraphics[width=15cm]{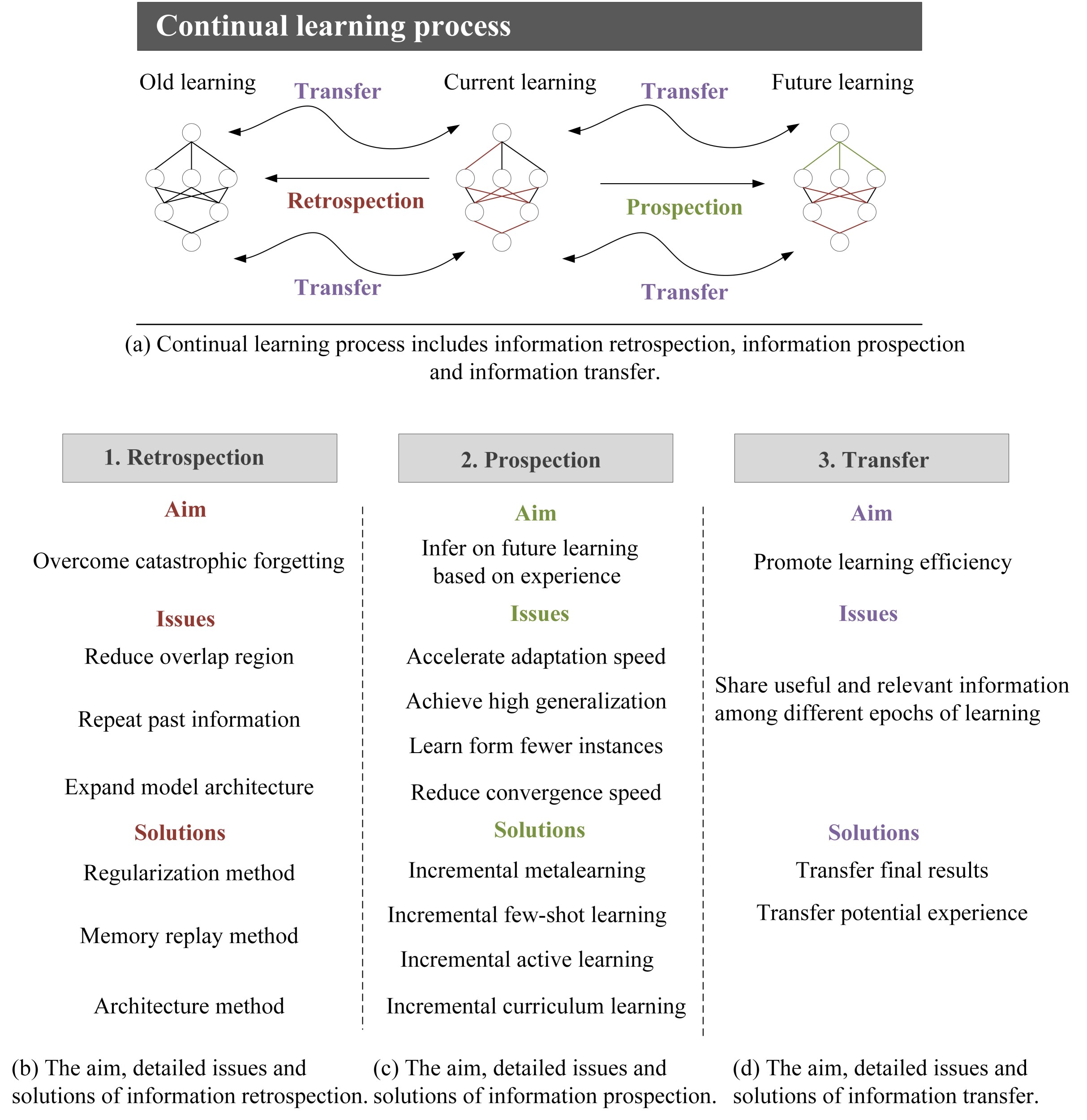}
\caption{The learning paradigm of an ideal CL agent.}
\label{img2}
\end{figure*}

As mentioned above, Stability Versus Plasticity mechanism highlights on both memory retention and future learning. Analogous to the mechanism, an AI agent is supposed to perform well in the three aspects in CL process, as shown in Figure \ref{img2} (a): i) Information retrospection: remembering previously learned information; ii) Information prospection: infer on new information continuously; iii) Information transfer: transferring useful information. Specifically:

\begin{itemize} \item \emph{Information retrospection:}  It is long-term memory of events, facts, knowledge and skills that have happened in the past \cite{ferbinteanu2003prospective}. One big challenge in information retrospection is the issue of catastrophic forgetting, the tendency of AI agents to completely and abruptly forget previously learned information upon learning new information \cite{mccloskey1989catastrophic}\cite{ratcliff1990connectionist}. This has motivated many ideas to address the problem such as reducing overlapped regions among different information, repeating past information, expanding network architecture to accommodate more information, and corresponding algorithms has also been proposed, as shown in Figure \ref{img2} (b). \item \emph{Information prospection:}  It is defined as the ability to remember to carry out intended actions in the future \cite{ellis2000prospective}. For AI agents, they are expected to deduce on future learning based on learned experience The ¡±smart¡± AI agents can accelerate learning speed, achieve high generalization, learn from limited data, and reduce convergence speed on future learning. Many learning paradigms are combined with CL such as incremental metalearning, incremental fewshot learning, incremental active learning and so on, to facilitate information prospection, as shown in Figure \ref{img2} (c). \item \emph{Information transfer:}  Information transfer aims to improve the learning in a task through the transfer of information from a related task \cite{torrey2009transfer}. The circulating and cross-utilized information can be exploited to facilitate future learning, and be supplemented to enhance the established knowledge system. Current CL learning problem are still now difficult to encapsulate and isolate into single domains or tasks. This has motivated many transfer learning techniques in CL algorithms, as shown in Figure \ref{img2} (d).
\end{itemize}

According to the taxonomy, this paper is organized as follows. Section 2 introduces the evaluation metrics in terms of information retrospection, information prospection and information transfer. Section 3 categorizes and describes the mainstream CL algorithms according to their contribution to the three aspects. Section 4 introduces the applications of CL. Section 5 presents recommendation regarding CL to promote a more extensive exploration in this field. Section 6 presents the concluding remarks.

\section{Evaluation metrics}
\begin{figure*}
\centering
\includegraphics[width=10cm]{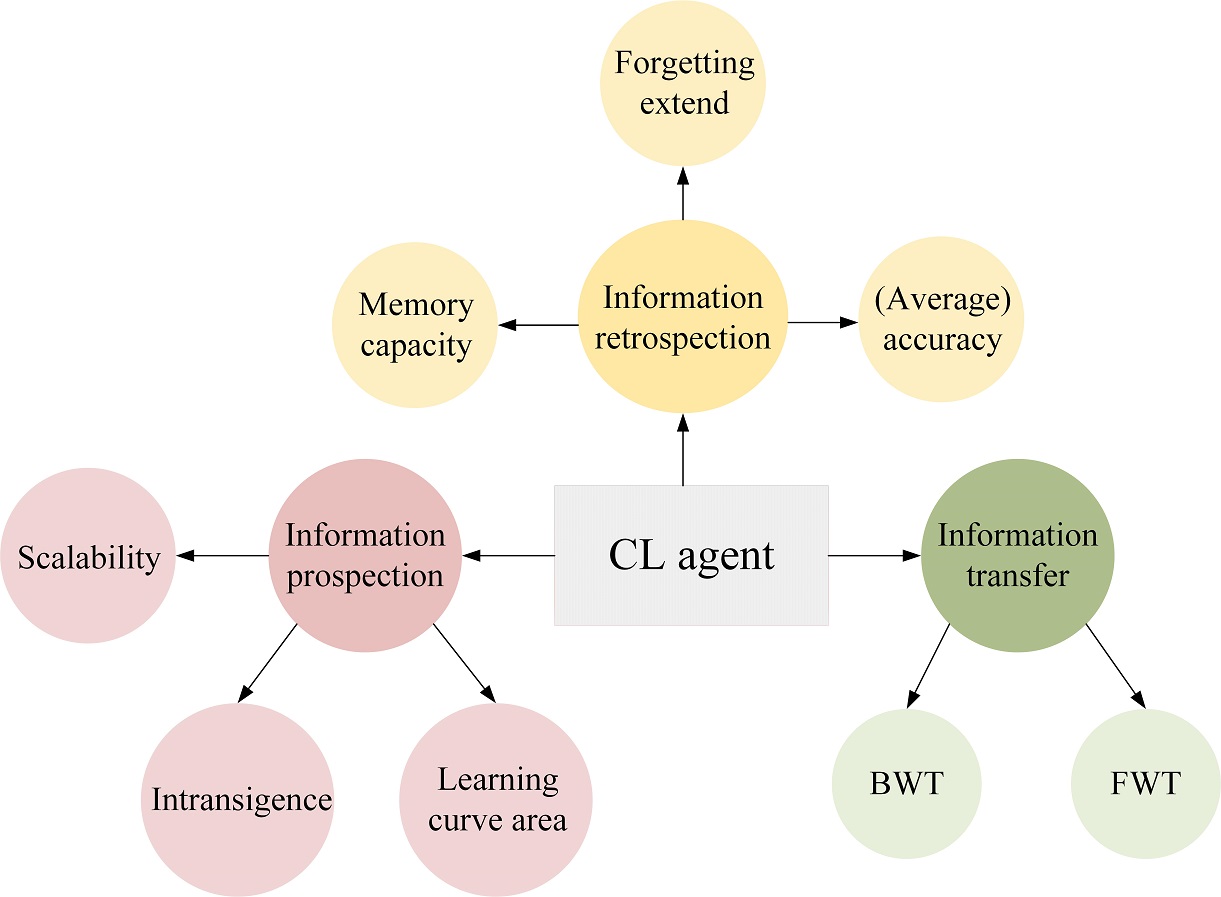}
\caption{Visualization of the evaluation metrics for a CL agent in terms of information retrospection, information prospection and information transfer.}
\label{img3}
\end{figure*}

The metrics to evaluate CL system lacks consensus, and the almost all metrics exclusively focus on the forgetting problem. In this section, a comprehensive summary on the performance of CL systems, in terms of information retrospection, information prospection and information transfer, is presented. The visualization of the evaluation metrics is presented in Figure \ref{img3}.

Information retrospection pertains to memory access to events and information that occurred or were learned in the past \cite{hunkin1995focal}. The loss of past information results in an abrupt deterioration in the performance pertaining to earlier knowledge after the learning of new information (catastrophic forgetting) in CNNs \cite{mccloskey1989catastrophic}. Information retrospection can be validated by the following three metrics.

\begin{itemize} \item \emph{Forgetting extend:} Forgetting is defined as the biggest knowledge difference for a specific task between the model has acquired continually till task \emph{k} and a previous one acquired before \emph{k} in the past \cite{chaudhry2018riemannian}. It represents the performance degradation on past tasks given its current state. For a classification problem, the forgetting for the \emph{j-th} task after the model has been incrementally trained up to task \emph{k (j \textless k)} is computed as Equation 1. \begin{equation}\begin{aligned}f_{j}^k&= \max\limits_{l\in\{1,\cdots,k-1\}}a_{l,j}-a_{k,j}\label{eq1}\end{aligned}\end{equation}
\item \emph{Memory capacity:} Memory capacity is defined as the number of learned tasks the model can remember at the current state. Recalling long sequence of previous tasks represents the outlasting memory retain ability. In fact, a simple fully connected model will show no forgetting on a two-task \cite{pfulb2019comprehensive}. Accordingly, testing CL agent on extremely long sequences of tasks is necessary. \item \emph{(Average) accuracy:} Accuracy pertains to the index of a model trained on training data and evaluated on test data \cite{lopez2017gradient}. In CL scenario, it is always computed for each previous task on the test data at the end of training on the current task. Average accuracy (ACC) is computed for all tasks at the end of a model¡¯s training for the current task continually, as shown in Equation 2. \begin{equation}\begin{aligned}A_{k}&= \frac{1}{k}\sum_{j=1}^{k}a_{k,j}\label{eq2}\end{aligned}\end{equation} Where $A_k$  is the ACC on all the task after the last task \emph{k} is learned; $a_{k,j}$ is the accuracy evaluated on the task \emph{j}, after the model has been trained with the final task \emph{k}.
\end{itemize}

Information prospection refers to the ability of a CL model to incorporate new information. The loss of memory prospection leads to a decreased ability to accumulate new information. The performance of memory prospection can be considered in terms of three aspects.

\begin{itemize} \item \emph{Learning Curve Area (LCA):} The LCA is a graphical representation of the rate of improvement in performing a task as a function of time or rate of change \cite{khan2014measuring}. This index represents the amount of time required by a model to acquire new information. Specifically, the learning curve in CL is regarded as convergence curve which describes an average performance at the \emph{j-th} mini-batch after the model has been trained for all the \emph{k} tasks, as shown in Equation 3. \begin{equation}\begin{aligned}Z_{b}&= \frac{1}{T}\sum_{k=1}^{T}a_{k,b,k}\label{eq3}\end{aligned}\end{equation} Where $a_{i,b,k}$  be the accuracy evaluated on the task \emph{j}, after the model has been trained with the \emph{b-th} mini-batch of task \emph{k}.
\item \emph{Intransigence:} Intransigence describes the inability of a model to learn new tasks \cite{chaudhry2018riemannian}. To quantify the intransigence on the \emph{k-th} task (denoted as $I_k$), the performance of the model trained on the \emph{k-th} task in the incremental manner (denoted as $a_{k,k}$) is compared with that of standard model which has access to all the datasets at all times (denoted as $a_k^\ast$), as shown in Equation 4. \begin{equation}\begin{aligned}I_{k}&= a_{k}^*-a_{k,k}\label{eq4}\end{aligned}\end{equation} \item \emph{Scalability:} Scalability is defined as the ability to ensure dynamic and efficient resource expansion or reuse when the weights are already saturated in the model \cite{rajasegaran2019adaptive}. To absorb as much new information as possible, it is necessary for a CL model to allocate a new set of weights to learn complimentary representations for a new task based on the sharing of common representations with old tasks.
\end{itemize}

Information transfer refers to the flexible knowledge transformation between retrospective and prospective information. In CNNs, the circulation of information can accelerate the learning of new information while reoptimizing the learning of old information. Memory transfer can be validated from two metrics, with the one to measure the impact that the new ones on the old ones, and vice versa.

\begin{itemize} \item \emph{Backward transfer (BWT)} \cite{lopez2017gradient}: The BWT indicates the influence that learning a new task has on the performance of former tasks. This index describes the extent to which a new task weakens the previous tasks. Learning new tasks may have two contrasting effects on the previous tasks: positive backward transfer (PBT) or negative backward transfer (NBT), which correspond to the promotion or degradation of the previous performance, respectively. The equation of BWT is in Equation 5. \begin{equation}\begin{aligned}BWT&= \frac{1}{k-1}\sum_{i=1}^{k-1}a_{k,j}-a_{i,i}\label{eq5}\end{aligned}\end{equation} \item \emph{Forward transfer (FWT)} \cite{lopez2017gradient}: FWT describes the influence that learning a current task has on the performance on a future task. In particular, a positive FWT occurs when the model can perform ¡°zero-shot¡± learning, likely by exploiting the structure specific to the task. The equation of BWT is in Equation 6. \begin{equation}\begin{aligned}FWT&= \frac{1}{k-1}\sum_{i=2}^{k}a_{i-1,i}-\overline{b_{i}}\label{eq6}\end{aligned}\end{equation} Where $\overline{b_{i}}$ is the vector of test accuracies for each task at random initialization.
\end{itemize}

\section{CL algorithms}
This section summarizes the mainstream CL algorithms referring to memory retrospection, memory prospection and memory transfer.

\subsection{Information retrospection}
Information retrospection is the most basic function of CL agents. The learned knowledge would likely be promptly forgotten, and new information would be difficult to acquire without it. Currently, many methods are available to preserve past information, and these methods can be categorized as regularization, memory and architecture methods.

\subsubsection{Regularization methods}
\begin{figure}[!t]
\centering
\includegraphics[width=2.5in]{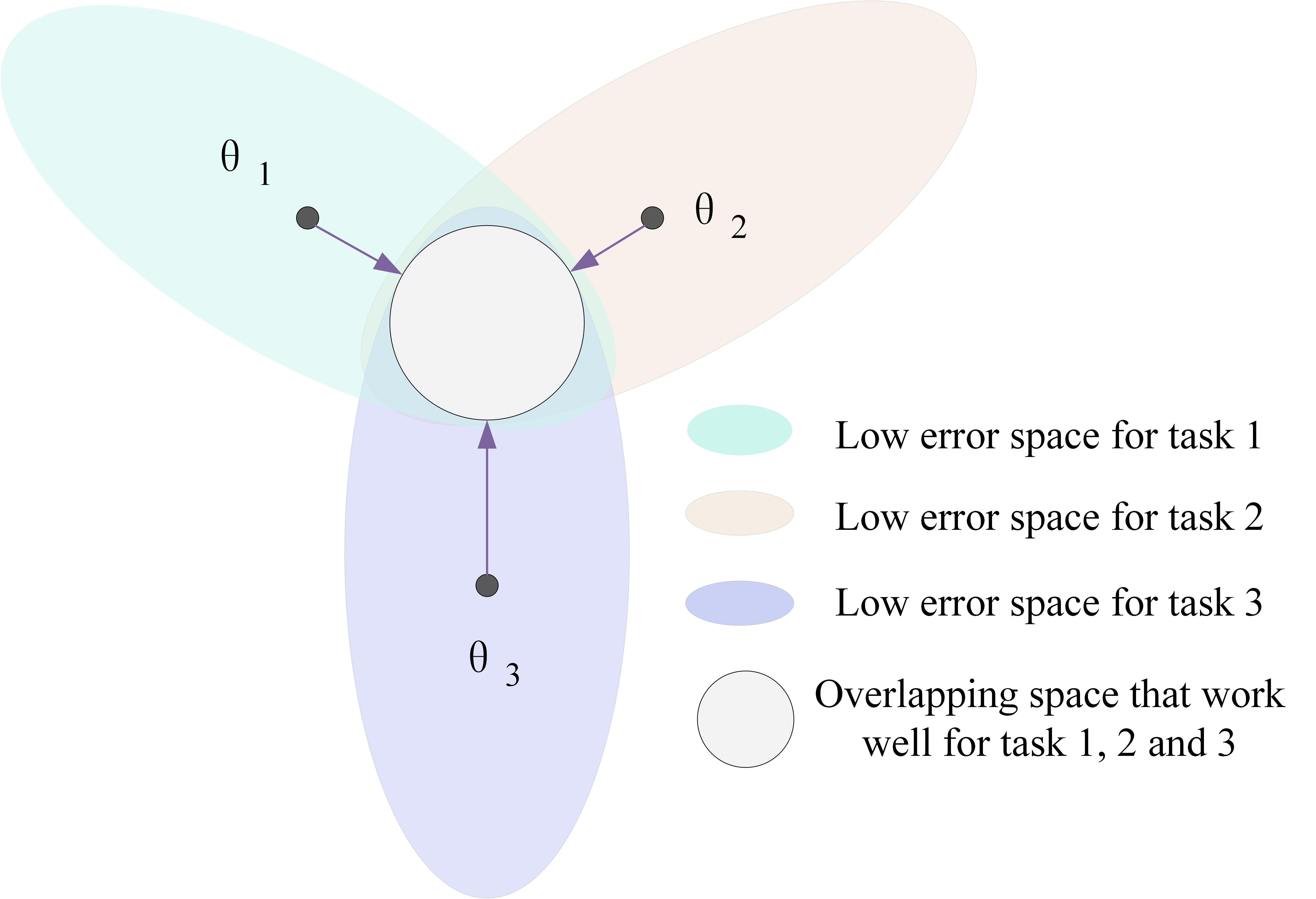}
\caption{Working mechanism of regularization method.}
\label{img4}
\end{figure}

\begin{figure*}
\centering
\includegraphics[width=15cm]{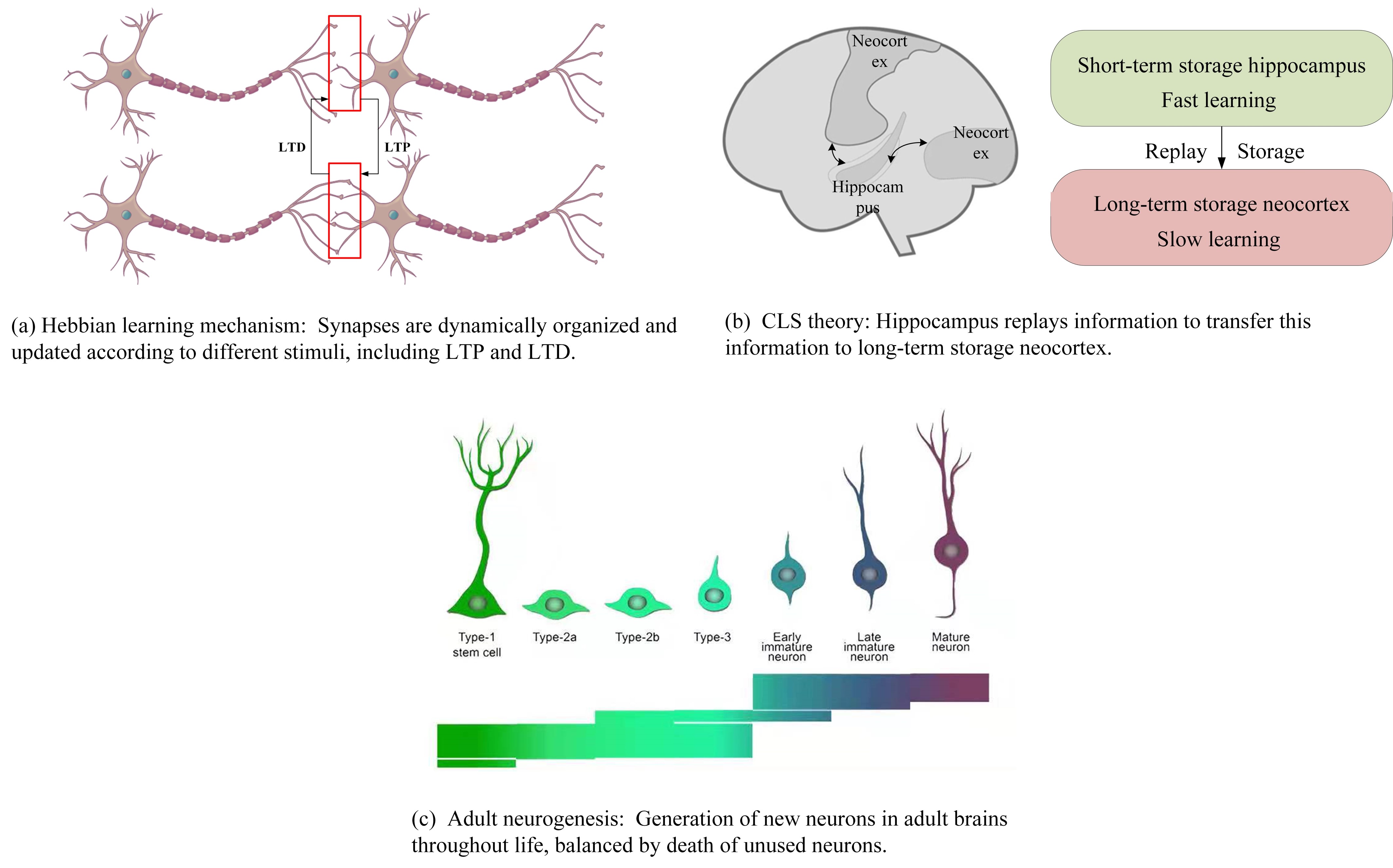}
\caption{The biological mechanisms for regularization method, memory replay method and architecture method.}
\label{img5}
\end{figure*}

Regularization methods limit the catastrophic forgetting phenomenon by imposing constraints as a regular term on the update of the weights in CNNs to help the network identify a set of weights that can provide proper mappings from each learned input to the output. The weights optimized using regularization methods may not be optimal for each pattern but include acceptable response values for all patterns, as shown in Figure \ref{img4}. Regularization methods are based on the concepts of Hebbian learning mechanism, according to which, synapses are dynamically organized and updated according to different stimuli \cite{hughes1958post}. High-frequency stimuli initiate long-term potentiation (LTP) in synapses, causing memory enhancement, whereas low-frequency stimuli result in long-term depression (LTD), causing memory degradation \cite{hughes1958post}, as shown in Figure \ref{img5} (a).

Elastic weight consolidation (EWC) \cite{kirkpatrick2017overcoming} is a pioneering and the most cited regularization method. EWC quantifies the importance of the learned parameters by estimating the Fisher information relative to the objective likelihood and preserves parameters with high importance values by restricting their drastic changes against new tasks. However, EWC assumes that the Fisher information matrix (FIM) \cite{liu2019quantum} is diagonal, which is unrealistic in the original parameter space. R-EWC \cite{liu2018rotate} enhances this diagonal assumption of EWC through a reparameterization strategy. This strategy rotates the parameter space by singular value decomposition (SVD) \cite{hinton2015distilling} such that the output of the forward pass is unchanged, but the FIM computed from the gradients during the backward pass is approximately diagonal. In this rotated parameter space, EWC can effectively optimize the new task. Although several methods similar to EWC have been proposed, these approaches calculate the parameter importance using different techniques. For example, according to the memory-aware synapsis (MAS) \cite{aljundi2018memory} technique, the change in an important weight can influence the output function of the model more significantly than changes in unimportant weights, and thus, this approach computes the importance of the weight by measuring the magnitude of the gradient of a parameter when it is perturbed. Compared with EWC, an apparent advantage of MAS is that it can update the model in an unsupervised and online manner by avoiding the need for labeled data. Synaptic intelligence (SI) \cite{zenke2017continual} computes the path integral of the gradient vector field as the weight importance along the entire learning trajectory. The most notable difference between SI and EWC is that SI computes the importance online and along the entire learning trajectory, whereas EWC computes the importance in a separate phase at the end of each task. Orthogonal weight modification (OWM) \cite{hu2018overcoming} optimizes the update direction of parameters in the direction orthogonal to all the previous input spaces. This strategy can avoid mutual interference among different tasks. To protect the most important weights, ABLL \cite{rannen2017encoder} leverages an autoencoder to capture the submanifold that contains the most informative features pertaining to a past task. When training for a new task, the features projected onto this submanifold are controlled to not be drastically updated. Rather than focusing on the weights in all layers of CNNs, LFL \cite{jung2016less} restricts the drastic changes in the learned parameters in the final hidden activations to preserve the previously learned input-output mappings and maintain the decision boundaries. IMM \cite{lee2017overcoming} progressively matches the Gaussian posterior distribution of the CNNs trained on the old and new tasks and uses various transfer learning techniques to render the Gaussian distribution smooth and reasonable.

Overall,the advantage of regularization methods is that it is computationally effective even with a small training period, low storage occupancy and low computational complexity. However, this approach is not scalable, resulting in a performance degradation when the number of classes increases gradually. In addition, the learning result is highly dependent on the relevance between tasks because the parameters for future tasks are bounded to the parameter regions generated by the past tasks.

\subsubsection{Memory replay methods}
The memory method consolidates old information by replaying them when models are updated for new tasks. The replayed old experience can be used to constrain the parameter updates such that the loss pertaining to previous tasks is not aggravated to retain connections for previously learned memories. The memory method originates from the hippocampal replay mechanism  in complementary learning systems (CLS) \cite{born2012system}, as shown in Figure \ref{img5} (b). Biological research has indicated that the hippocampus in rodents replays information that they experience in the daytime during sleep or rest \cite{o1978hippocampus} to transfer this information to long-term storage.

\begin{figure}[!t]
\centering
\includegraphics[width=3.3in]{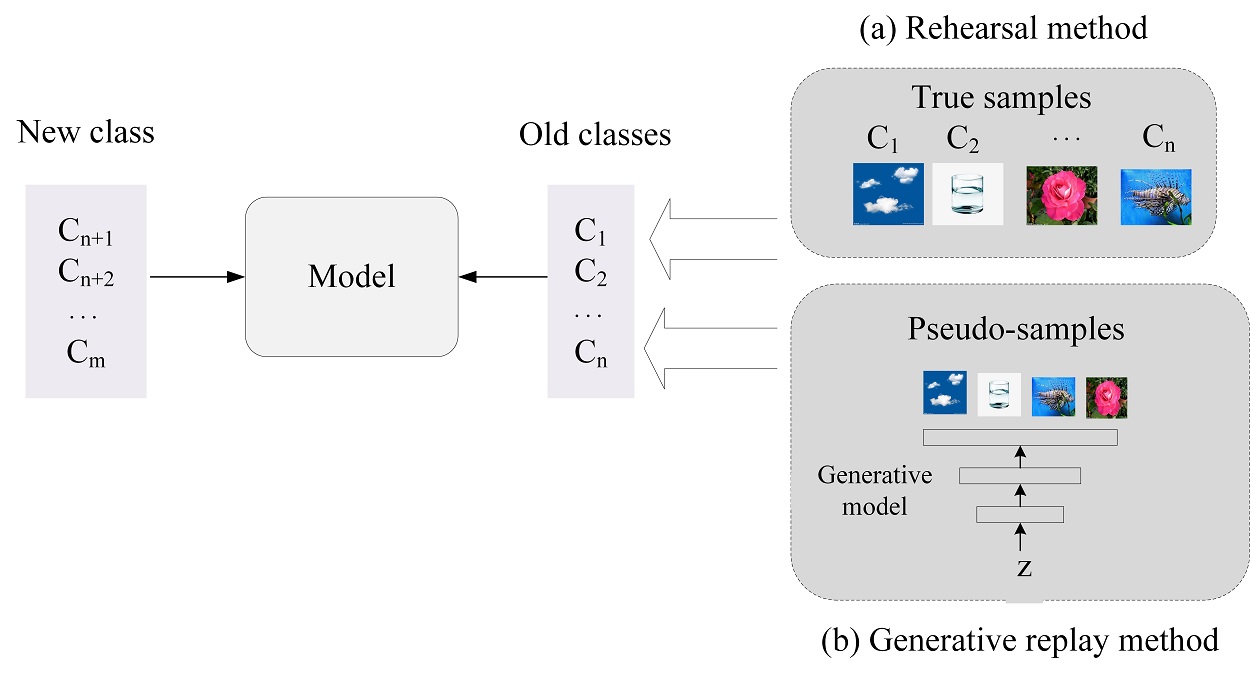}
\caption{Working mechanism of the memory method [39], including the rehearsal and generative replay methods.}
\label{img6}
\end{figure}

Memory replay methods can be broadly categorized into two types of methods, namely, rehearsal and generative replay methods, according to whether the replayed past data are real data.

The rehearsal strategy stores data from previously learned tasks with a memory buffer and interleaves them with the training data of the current task to jointly train the model, as shown in Figure \ref{img6} (a). Incremental classifier and representation learning (iCaRL) \cite{rebuffi2017icarl} is a representative approach. iCaRL employs an exemplar set including the most representative samples of each previous class. These samples can most accurately approximate the average feature vector over all training examples. Next, the approach applies the nearest mean classifier \cite{veenman2005nearest} strategy to predict a label for a new image based on the exemplar set. iCaRL can incrementally learn many classes over a long period; however, it updates the exemplar set and classifier independently. End-to-end incremental learning (EEIL) \cite{castro2018end} overcomes the limitation of iCaRL by jointly learning the classifier and features. The approach achieves end-to-end learning by formulating an integrated cross-distilled loss based on the distillation loss to extract representative samples from old data and using the cross-entropy loss to simultaneously learn new classes. Subsequently, the imbalance problem between previous and new data is considered. The unified classifier (UC) \cite{hou2019learning} adopts a cosine normalization-based classifier to eliminate the significant difference in the bias and weights between old and new tasks. Subsequently, the approach employs a less-forget constraint strategy to ensure that the features of the old samples in the new and old models do not significantly differ. According to large-scale incremental learning (LSIL) \cite{wu2019large}, the class imbalance problem causes the classifier to classify an image into the category with a larger amount of data; therefore, the approach introduces a bias correction (BiC) layer to correct the bias regarding the output logits for the new classes. In the incremental learning with dual memory (IL2M) \cite{belouadah2019il2m} approach, a dual memory is introduced to alleviate the negative effect of the imbalance problem. The first memory stores the exemplar images of past classes. The second memory stores the initial class statistics in a highly compact format as it is assumed that the initially learned classes are best modeled. Experimental results show that the initial class statistics stored in the second memory can help the model overcome the problem of imbalanced datasets in past classes and rectify the associated prediction scores. In addition to the sample imbalance problem, certain approaches have been developed to optimize the memory storage utilization. For example, MECIL \cite{zhao2021memory} aims to optimize the exemplar set management. This approach retains low-fidelity exemplar samples rather than the original high-fidelity samples in the memory buffer to ensure that the limited memory can store more exemplars. Certain other approaches attempt to enhance the stored items of previous tasks. For example, instead of directly rehearsing the stored examples, the gradient episodic memory model (GEM) \cite{lopez2017gradient} stores gradients of the previous task to define inequality constraints regarding the loss to ensure that the loss does not increase with respect to that in previous tasks.

However, the rehearsal strategy relies on stored data, which is undesirable for several reasons. First, data storage is not always possible in practice due to safety or privacy concerns. Second, the approach is difficult to scale up to address problems involving many tasks. Finally, the rehearsal method is questionable in terms of the neuroscience perspective because the brain does not directly store data, such as all pixels of an image. As an alternative, generative replay (GR) has been proposed to rehearse past data without having access to them. In contrast to restoring the exact samples in old tasks, this approach uses a separate generative model to generate a pseudosample of old tasks, as shown in Figure \ref{img6} (b). Deep generative replay (DGR) \cite{shin2017continual} introduces generative adversarial networks (GANs) \cite{creswell2018generative} to mimic past data. The generated pseudodata and their responses pertaining to the past model are paired to represent old tasks, and the data are interleaved with new data to update the model. DGR achieves promising continual learning results; however, the generator must be repeatedly trained using a mix of samples synthesized for previous categories and real samples of new classes. Therefore, certain researchers attempted to reduce the computational overhead for the replayed data based on DGR. For example, the deep generative memory (DGM) \cite{ostapenko2019learning} approach eliminates the reuse of previous knowledge by introducing learnable connection plasticity for the generator. The approach designs task-specific binary sparse masks for the learnable units of the generator weights. The gradients of weights in each layer of the generator are multiplied by the reverse of the cumulated mask to prevent the overwriting of previous knowledge. Rather than replaying the real samples, BIR \cite{van2020brain} replays the internal or hidden representations of past tasks. The replayed representation is generated by the context modulated feedback connection of the network. GRFC \cite{van2018generative} reduces the computational cost by integrating the generative model with the main model through generative feedback connections. In addition, because combining DGR with a distillation strategy can enhance the performance, the approach labels the input data of the current task by using the model trained for the previous tasks as soft targets and uses the resulting input-target pairs as the pseudodata. DGM, BIR and GRFC are more scalable than DGR in the case of complicated problems involving many tasks or complex inputs. It has been highlighted that reliable data help retain the corresponding past information, and thus, certain researchers focused on enhancing the quality of the replayed data. For example, MeRGAN \cite{wu2018memory} uses a conditional GAN, in which the category is used as an input to guide the replay process, thereby avoiding less biased sampling for past categories and generating more reliable past data. CloGAN \cite{rios2018closed} uses an auxiliary classifier to filter a portion of distorted replays to block inferior images from entering the training loop.

In general, the memory method achieves satisfactory results in continual learning, and many inherent issues, such as the efficacy of memory buffers, have been considered with the research progress. The memory replay method is robust against catastrophic forgetting when the relevant experience is elaborately selected. However, an inherent problem is the long training time when training models with mixed data. In addition, the problem of imbalanced samples when rehearing or replaying past data has not been fully solved.

\subsubsection{Architecture methods}
\begin{figure}[!t]
\centering
\includegraphics[width=2.5in]{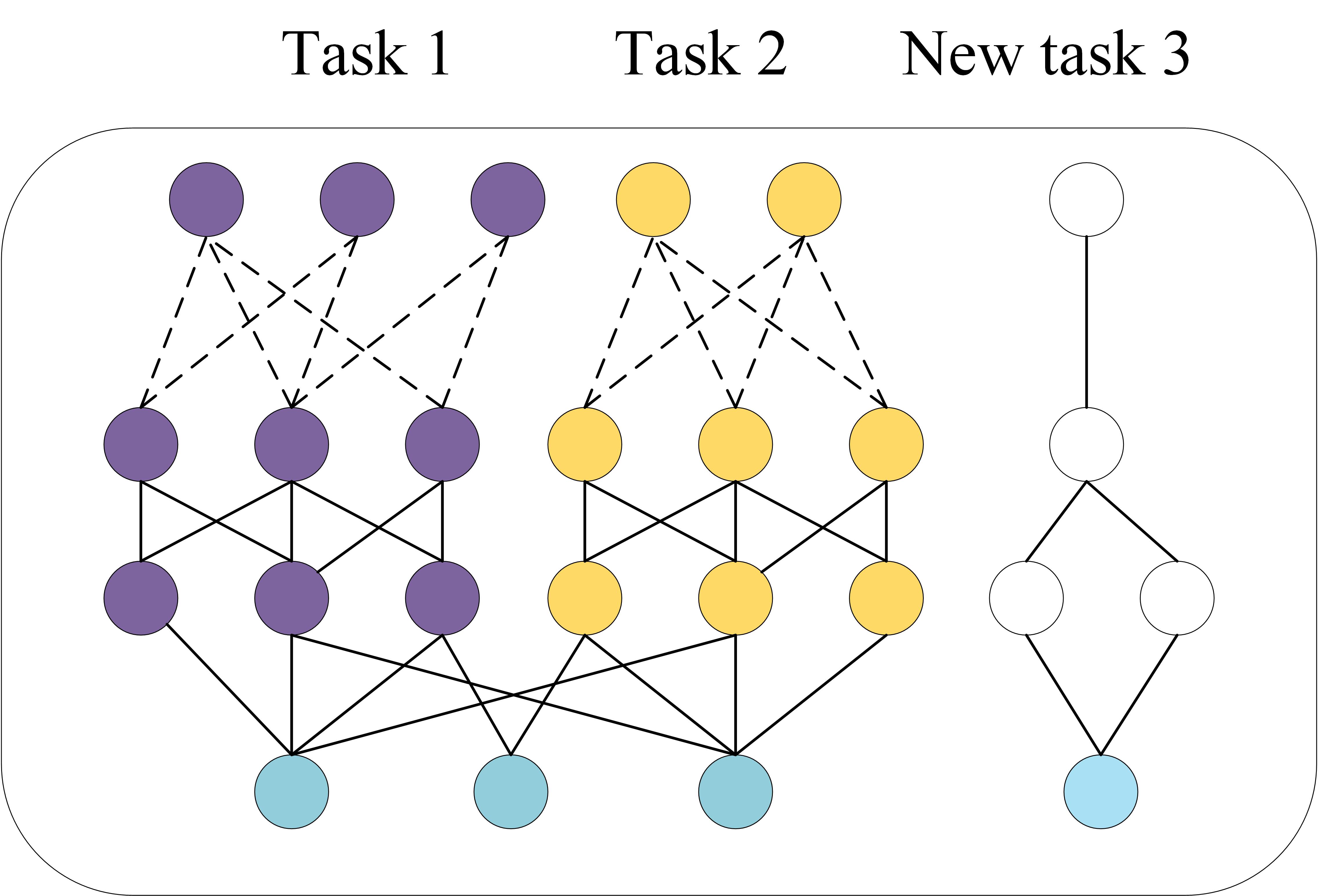}
\caption{Working mechanism of the architecture method.}
\label{img7}
\end{figure}

The architectural strategy realizes memory retrospection by flexibly expanding model architectures to accommodate more information while preserving part of the architectures trained for previous tasks. Different subsets of model parameters are assigned to different tasks in architecture method, as shown in Figure \ref{img7}. Research on the architecture method is motivated by the adult neurogenesis theory \cite{zhao2008mechanisms}, as shown in Figure \ref{img5} (c). Adult neurogenesis is the formation of functional, mature neurons from neural stem cells in specific brain regions in adults. In these regions, new neurons are generated throughout life and integrated into established neuronal circuits \cite{zhao2008mechanisms}.

There are two main categories of architectural strategies, namely, fixed networks and dynamic networks, based on whether the model structure is expanded.
\begin{table*}[!t]
\renewcommand{\arraystretch}{1.3}
\caption{Advantages and disadvantages of the regularization methods, memory replay methods and architecture methods}
\label{table1}
\centering
\begin{tabular}{|p{2.9cm}|p{6cm}|p{5cm}|}
\hline
CL methods & Advantage & Disadvantage\\
\hline
\multirow{3}{*}{Regularization method} & Small training period  & \multirow{2}{*}{Low scalability} \\

  & Low storage occupancy &  \multirow{2}{*}{Mediocre performance}\\
  &  Low computational complexity &\\
\hline
\multirow{3}{*}{Memory method} & \multirow{2}{*}{Promising performance} & Large training period (Rehearsal)\\
 & \multirow{2}{*}{Small training period (Generative replay)} & High storage occupancy\\
 & & Sample imbalance\\
\hline
\multirow{3}{*}{Architecture method} & \multirow{3}{*} {Low storage occupancy} & Low scalability \\
 & & Mediocre performance\\
  & & High computational complexity\\
\hline
\end{tabular}
\end{table*}
The fixed network strategy only allows inner network adjustments, such as changes in the weights and activations. PackNet \cite{mallya2018packnet} is a typical fixed network technique. This approach employs a weight-based pruning technique \cite{han2016dsd} to release redundant parameters across all layers after the training for a task. The remaining parameters retain the information for old tasks, and the released parameters are updated for new tasks. PathNet \cite{fernando2017pathnet} employs evolutionary strategies \cite{shukla2015comparative} to select pathways that determine the parameters in the network to be retained or updated. The approach fixes the parameters along a path learned on the old task and re-evolves a new population of paths for the new task.

In contrast to enforcing a predefined architecture, a dynamic network allows the weights to be added or removed according to the future tasks. PNN \cite{rusu2016progressive} expands the architecture by allocating novel subnetworks with a fixed capacity to be trained using the new information. OIFL \cite{zhou2012online} is based on adaptive feature learning that adds features for samples with a high loss and subsequently merges the similar features to prevent based on a denoising autoencoder. Instead of enforcing a predefined architecture, Adanet \cite{cortes2017adanet} automatically evolves to learn network structures. Starting from a simple linear model, Adanet adds more units and additional layers, as necessary. The added units are carefully selected and penalized according to rigorous estimates from several theories of statistical learning \cite{koltchinskii2002empirical}\cite{kotani1997structural} to seek high-quality models with minimal expert intervention. EDIL \cite{xiao2014error} employs a training algorithm that grows the network capacity in a hierarchical manner. Similar classes are grouped and self-organized into levels, with models cloned from previous classes and trained in parallel to accelerate the training. The neurogenesis deep learning (NDL) model \cite{draelos2017neurogenesis} adds new neurons to the autoencoder and employs intrinsic replay to reconstruct old samples and preserves their learned weights. Following this idea, a dynamically expandable network (DEN) selectively \cite{yoon2017lifelong} retrains the old network and expands its capacity when necessary, thereby dynamically deciding the optimal capacity of the network being trained online. RPSnet \cite{rajasegaran2019random} adopts a random path selection algorithm that progressively chooses optimal paths for the new tasks while encouraging parameter sharing and reuse.

Generally, architecture methods exhibit a low storage occupancy since they do not store any extra samples or parameters. However, such methods are not scalable because the network architecture may not be sufficient or be randomly expanded as the number of tasks increases. In addition, the computational complexity is slightly high because the approaches recompute the pathways in the network for each new task.

In conclusion, the three types of methods to achieve memory retrospection are generally effective and exhibit characteristic merits and unavoidable defects, as summarized in Table \ref{table1}. These algorithms are summarized in Table \ref{table2}.

\begin{table*}[!t]
\renewcommand{\arraystretch}{1.4}
\caption{Summary on regularization methods, memory replay methods and architecture methods.}
\label{table2}
\centering
\begin{tabular}{|c|c|c|c|}
\hline
Method & Algorithm & Keypoint & Reference\\
\hline
\multirow{8}{*}{Regularization method} & EWC & Measure weight importance & [24]\\
 & R-EWC & Rotates parameter space & [26]\\
 & MAS & Label free & [28]\\
 & SI & Online training & [29]\\
 & OWM & Orthogonal weight modification & [30]\\
 & ABLL & Capture informative feature & [31]\\
 & LFL & Focus on hidden layer & [32]\\
 & IMM & Bayesian moment matching & [33]\\
\hline
\multirow{7}{*}{Memory replay method - Rehearsal} & iCaRL & Representation Learning, Nearest-Mean-of-Exemplars Classification & [36]\\
& EEIL & Cross-distilled loss function  & [38] \\
& UC & Cosine Normalization  & [39] \\
&  LSIL &  Store initial class statistics & [40] \\
&  IL2M &  Bias correction method  & [41]\\
& MECIL  & Low-fidelity exemplar  & [42]\\
&  GEM & Store previous gradient  & [20]\\
\hline
\multirow{6}{*}{Memory replay method - Generative replay} & DGR & Introduces GAN & [43]\\
& DGM & Task-specific binary mask & [45]\\
& BIR & Replay hidden representation & [46]\\
& GRFC & Distillation and  Replay-through-Feedback & [47]\\
& MeRGAN & Use conditional GAN & [48] \\
& CloGAN  & Filter out distorted replay & [49]\\
\hline
\multirow{9}{*}{Architecture method} & PackNet &  Iterative pruning and re-training technique & [51]\\
& PathNet & Computationally cost-effective & [53]\\
& PNN & Expand network architecture & [55]\\
& OIFL & Optimize feature set & [56]\\
& Adanet & Auto structure search  & [57] \\
& EDIL  & Expand network hierarchically  & [60] \\
& NDL & Applied in AE & [61] \\
& DEN & Selective retraining & [62]\\
& RPSnet  &  Random search & [63]\\
\hline
\end{tabular}
\end{table*}

\subsection{Information prospection}
Preserving learned information is important but not sufficient in the CL scenario. The abovementioned methods can address forgetting aspects; however, they may not be able to facilitate future learning. Constant incorporation and efficient acquisition of new information are necessary in a long-term learning process. Only a few CL approaches have been specifically designed to leverage memory prospection, although certain interdisciplinary fields have combined CL with other learning paradigms, such as active learning and few-shot learning have witnessed remarkable enhancements in the promotion of memory prospection.

\subsubsection{Incremental metalearning}
Metalearning \cite{hospedales2020meta} is most commonly understood as learning to learn and refers to the process of enhancing a learning algorithm over multiple learning episodes. This approach uses the metadata regarding past experiences, such as hyperparameters, to promptly learn new experiences. The aim is to progressively increase learning efficiency while learning an increasing number of tasks, which is also the objective of CL in memory prospection. An increasing number of researchers have designed or adopted metatraining strategies such as OML \cite{javed2019meta}, MAML \cite{finn2017model} and Reptile \cite{nichol2018reptile} to enhance the memory prospection in CL. These metatraining algorithms can finetune the model optimization to enable prompt adaptation to new tasks.

An example of applying OML in continual metalearning pertains to the work of Javed and White \cite{javed2019meta}. To address the continual learning prediction problem, the model is divided into two parts, specifically, the RLN that learns the features of inputs by representation learning, and the PLN that predicts a class for inputs based on the learned representations. In the PLN, the parameters are updated in a metalearning manner guided by online aware metalearning (OML). OML can learn a highly sparse and well-distributed representation by exploiting the large capacity of the representation. This aspect can help reduce forgetting because each update changes only a small number of weights, which in turn only affect a small number of inputs. Experimental results show that the OML is robust to interference under online updates and promotes future learning.

In addition to OML, MAML \cite{finn2017model} and its variant FOMAML \cite{finn2017model} have been utilized. For example, Gupta et al. \cite{gupta2020maml} optimize the OML objective in an online way through a multistep MAML procedure. The authors indicate that the gradient alignment among old tasks does not degrade while a new task is learned in OML; therefore, it is necessary to avoid repeated optimization of the inter-task alignment between old tasks to enable acceleration. Considering this aspect, the authors propose C-MAML to focus only on aligning the gradients of the current task and average gradient of the previous tasks. To prevent forgetting, the authors adopt La-MAML, in which the learning rates of the inner loop are clipped to positive values to avoid gradient ascension and interfering parameter updates, thereby mitigating catastrophic forgetting. SeqFOMAML \cite{spigler2019meta} uses FOMAML to promptly learn tasks in the sequence and discourages interference between tasks. This approach first learns a prior by MAML to initialize the model parameters and relies on the plain stochastic gradient descent in the continual learning process over a sequence of tasks. Experimental results show that SeqFOMAML exhibits a superior performance for metageneralization to longer sequences. MAML is extended to Continual-MAML in the OSAKA framework \cite{caccia2020online}. Continual-MAML has two stages, namely, the pretraining phase, which initializes the model with parameters learned by the MAML, and the CL phase, which adapts the learned parameter initialization to solve new tasks. When a change in the distribution is detected, Continual-MAML adds the new knowledge into the learned initialization. With the introduction of Continual-MAML, OSAKA can rapidly solve new tasks and remember old tasks.

Furthermore, Reptile \cite{nichol2018reptile} has gained attention. For example, Riemer et al. propose meta-experience replay (MER), which can facilitate continual learning by maximizing useful transfer and minimizing interference. The objective is aimed at encouraging the network to share parameters when the gradient directions align and keep parameters separate when interference is caused by the gradients in opposite directions. To this end, this approach interleaves new inputs with the examples sampled from the replay buffer and later updates the parameters in the model based on Reptile by using these mixed data. Reptile can maximize the inner product between gradients of different minibatches from the same task, corresponding to enhanced generalization. By combining experience replay with optimization-based metalearning Reptile, the updated parameters are more likely to be transferred and less likely to incur interference with respect to past examples in the MER.

Instead of applying a specific metalearning algorithm, He et al. \cite{he2019task} propose a framework in which a wide range of meta-algorithms can be embedded, such as MAML, LEO \cite{rusu2018meta}, CAVIA \cite{zintgraf2019fast} and CNP \cite{he2019task}. The aim of the framework is to exploit metalearning to accelerate the recovery of lost performance rather than focusing on remembering previous tasks, aided by the explicit inference of the current observed task. The authors design a metaframework known as What \& How functions, which consists of an encoder or task inference network that predicts the current task representation based on the context data, and a decoder that maps the task representation to a task specific model. By using the continual learning strategy BGD \cite{zeno2018bayesian}, the metaparameters in the framework can be updated continually to learn a sequence of tasks.

In general, metalearning enables prompt adaptation to new tasks in CL due to its powerful generalization ability. An increasing number of classic metalearning algorithms have been utilized in CL, and novel metalearning strategies specifically designed for CL are being gradually examined.

\subsubsection{Incremental few-shot learning}
Few-shot learning \cite{fei2006one} refers to the ability to learn to recognize new concepts based on only a few samples. The research on few-shot learning mainly focuses on achieving a high generalization over new tasks with a limited number of training data. Historically, most methods pertaining to CL are based on the assumption that relatively large batches of training data are available. However, this assumption substantially limits the scalability of these approaches to the open-ended accommodation of novel classes with limited training data. Incremental few-shot learning is an promising approach to address this dilemma in the CL scenario. The key aspect in incremental few-shot learning is to prevent overfitting for new classes with only a few training samples on the basis of preserving old information.

To address the overfitting problem, few-shot learning adopts episodic training \cite{vinyals2016matching}. Specifically, tasks are sampled from support examples on which the learner is guided to learn. Subsequently, the learner is evaluated on query examples with an objective function to determine how well the learner can guide the learner to generalize to new tasks. Finally, the learner is optimized using the objective function. Through repetitive episodic training, the learner can gradually generalize over few-shot tasks. This training method has been widely incorporated in incremental few-shot learning.

Open-ended CentreNet (ONCE) \cite{perez2020incremental} is designed to incrementally detect novel class objects with few examples. First, a feature extractor is trained with abundant base class training data and frozen to preserve the learned information. Subsequently, a generator that can synthesize class-specific codes for novel classes is trained, and finally, an object locator is trained to detect objects. The class-specific codes are specifically designed for each new task, and the information does not change the original weights in the model; thus, the forgetting problem can be addressed. When training the generator, ONCE adopts an episodic metalearning strategy in which a support set is constructed to train the model to learn and a query set is used to train the model to generalize. Cheraghian et al. \cite{cheraghian2021semantic} propose a semantic-specific information method for incremental few-shot classification tasks. Specifically, given an image as input, the semantic word vectors are estimated, and the similarity of the predicted word vectors with the word vectors from the set of possible class labels is measured to obtain the final class. The semantic information for novel classes is generated by the combination of multiple embedding features and global features trained via episodic training. In this way, the model can attain a high generalization when classifying both base and novel classes. Xiang et al. \cite{xiang2019incremental} adopt incremental few-shot learning for pedestrian attribute recognition. Specifically, the authors design APGM that extracts the multiple-attribute information from feature embedding and produces classification weights for the novel attributes. Subsequently, the approach samples \emph{N}-way \emph{K}-shot tasks from the base class as the support set and treats the sampled attributes from the support set as fake novel attributes. Next, the classification weights for fake novel attributes are generated under the guidance of APGM. Finally, the classification performances of the classification weights on the query set are evaluated, and the gradient is backpropagated to update the APGM. Since the APGM module does not interfere with the base model, the performances over the base data are not hampered. CBCL \cite{ayub2019cbcl} involves the generation of a set of concepts in the form of centroids for each class using a clustering approach known as Agg-Var clustering. After generating the centroids, to predict the label of a test image, the distance of the feature vector of the test image to the \emph{n} closest centroids is used. Since CBCL stores the centroids for each class independent of the other classes, the decrease in the overall classification accuracy is not catastrophic when new classes are learned. CBCL has been implemented to incremental robotic object detection tasks in which a robot learns different categories of objects from only a few examples. Experimental results show that even after learning 22 different categories of objects, the robot can correctly recognize previously learned objects at an accuracy of approximately 90\%.

Incremental few-shot represents an emerging research direction, and it has been widely examined since the 2020 year \cite{ayub2020tell}. Current incremental few-shot algorithms have considerably enhanced memory prospection by exploring and exploiting the structural information among different tasks.

\begin{table*}[!t]
\renewcommand{\arraystretch}{1.4}
\caption{Summary on incremental metalearning, incremental few-shot learning, incremental active learning.}
\label{table3}
\centering
\begin{tabular}{|c|c|c|c|}
\hline
Method & Algorithm & Keypoint & Reference\\
\hline
\multirow{6}{*}{Incremental metalearning} & Javed and White & RLN and PLN & [65]\\
 & Gupta et al & Avoid repeated optimization & [68]\\
 & SeqFOMAML  & learn a prior initialization by MAML g & [69]\\
 & Continual-MAML & OSAKA framework & [70]\\
 & MER & Update parameters by Reptile  & [67]\\
 & He et al & Universial metaframework  & [71]\\
\hline
\multirow{4}{*}{Incremental few-shot learning} & ONCE & Meta-learned generator network & [77]\\
& Cheraghian et al & Meta-learned semantic information detector  & [78] \\
& Xiang  et al & Meta-learned feature abstracter  & [79] \\
&  CBCL  &  Generate the centroids & [80] \\
\hline
\multirow{3}{*}{Incremental active learning} & Ahmed et al & Two-level sample selection strategy & [87]\\
& MEN & Modified entropy Learning strategy & [88]\\
& Lin et al & Independent metrics & [90]\\
& Zhu et al & Bayesian network & [92]\\
\hline
\end{tabular}
\end{table*}

\subsubsection{Incremental active learning}
Active learning algorithms are used in situations in which abundant unlabeled data are available but manual labeling is expensive. Active learning can interactively query a user (or another information source) to label new data points \cite{settles2009active}. The key concept is that an algorithm can achieve a higher accuracy with fewer training instances if it is allowed to choose the data from which it learns.

Incremental active learning draws on this idea to enhance memory prospection in CL. When active learning is performed, CL can be realized without learning each instance in future learning. The most useful data for current learning can be selected based on query strategies in active learning, for instance, by using the uncertainty sampling strategy \cite{lewis1994sequential}, expected model change strategy \cite{settles2007multiple} and variance reduction and Fisher information ratio strategy \cite{cohn1996active}. In this way, informative and critical data are preserved, whereas redundant and useless data are removed, thereby increasing the efficiency of future learning. To this end, designing appropriate query selection strategies for each new specific scene while preserving old scenes is important in incremental active learning, as shown in Figure 10. The subsequent section describes the ways in which different query selection strategies are applied to improve CL.

Ahmed et al. \cite{qureshi2020active} present a two-level sample selection strategy that can enable motion planning networks (MPNets) \cite{qureshi2019motion} to learn from streaming data. To address the forgetting problem, the GEM method and replayed memory are employed to ensure lifelong learning. When updating the episodic memory in GEM and replayed memory, the approach first collects the training data by actively asking for demonstrations on problems in which MPNet fails to find a path. Second, the approach prunes the expert demonstrations to fill the episodic memory to enable the approximation of the global distribution from the streaming data. The online MPNet is successfully applied to robotic motion planning and navigation and exhibits comparable planning performance in tasks with approximately 80\% less data than traditional approaches.

Shi et al. \cite{shi2020incremental} introduce a novel active learning strategy known as modified entropy (MEN) to achieve incremental atrial fibrillation detection. Transfer learning is used to address forgetting. While continuously updating the model, the MEN selects the most useful information from massive medical data as the learning set. Specifically, the features and uncertainty of the predicted results are integrated to select the informative samples. Traditionally, samples with a high entropy are believed to be abundant in information. MEN assumes that samples with a high RR intervals \cite{malik2002relation} series but low entropy are also informative. Next, the model is continuously finetuned using the most informative samples selected by MEN.

Lin et al \cite{lin2020active} propose a sampling strategy specifically designed for the semantic segmentation of large-scale ALS point clouds \cite{melzer2004extraction} in urban areas. To identify the most informative parts of a point cloud for model training, this approach quantitatively assesses both data-dependent and model-dependent uncertainties by using three independent metrics. The data-dependent uncertainty is estimated through the point entropy and segment entropy. Segment entropy considers the interactions among neighboring points. The model-dependent uncertainty is estimated by mutual information, and the disagreements produced by different model parameters are evaluated. Experimental results demonstrate that all three metrics can help select informative point clouds that can be generalized to point clouds through terrestrial mobile laser scanners or indoor scenes.

Furthermore, Zhu et al. \cite{zhu2013mathematical} introduce an active and dynamic selection method for crop disease diagnosis algorithms. The authors selected a subset of symptoms that are most relevant to the diagnosis of crop diseases based on a Bayesian network. In this approach, the symptoms are represented by the nodes of Bayesian networks, and another node pertaining to ¡°diseases¡± is introduced, with each value of this node representing one crop disease. A Markov blanket \cite{tsamardinos2003algorithms} is used to obtain the symptom nodes that most significantly influence the disease. Experiments show that the proposed method is suitable for diagnosing not only crop diseases but also other diseases by extending the values of the disease variable and symptom variables to other disease situations.

Overall, incremental active learning algorithms can increase the efficiency of grasping new information in CL. Future work can be aimed at developing additional real-time incremental active learning algorithms because the current methods are slightly time-consuming in the context of the query strategy.

\subsubsection{Incremental curriculum learning}
In curriculum learning, the models first consider easy examples of a task and the task difficulty gradually increase over a sequence of tasks \cite{bengio2009curriculum}. The objective of curriculum learning is to solve the last task, whereas the objective of CL is to be able to solve all tasks. In fact, the aim of finding a globally optimum CL can be achieved by gradual optimization in curriculum learning. Specifically, curriculum learning first optimizes a smoothed objective pertaining to easy examples, gradually adjusts the objective while maintaining the loss at a local minimum, and finally obtains the parameters that are globally optimal for all data.

The core of adopting curriculum learning in CL is to rearrange the sequence of examples according to their complexity. Intuitively, selecting a simpler example and building on this example to learn increasingly sophisticated representations are less time consuming than directly attempting to manage an unlearnable example when encountering a batch of new data. In addition, starting with easy samples can help the model avoid ¡°noisy¡± data, which reduces the convergence speed. Although difficult examples are more informative than easy examples, they are likely not useful because they confuse the learner rather than helping it establish the correct location of the decision surface \cite{bengio2009curriculum}. In addition, experiments demonstrate that the generalization error pertaining to training on simple data is lower than that pertaining to training on randomly selected data. Therefore, the curriculum strategy can increase the learning efficiency, enhance the generalization ability and increase the convergence speed for acquiring new knowledge in CL.

According to this analysis, curriculum learning can considerably enhance the memory prospection of CL models if the learning sequence for the new data is ordered. Regretfully, only a few incremental curriculum learning algorithms have been developed, and the potential advantages of curriculum learning in the context of CL should be explored in the future.

Generally, efficiently grasping new knowledge is important for a CL model to manage situations involving unknown tasks. Multiple learning paradigms are demonstrated to be effective in leveraging memory prospection when they are incorporated into CL. The LCA, intransigence and scalability are superior to those of plain CL algorithms. These new learning methods are evaluated according to their performance and potential reported in the original papers, as shown in Table \ref{table3}.

\subsection{Information transfer}
The actual world is structured, and most components are correlated; new situations look confusingly similar to old ones. In such a scenario, past information can be drawn on to facilitate future learning, and new information can be supplemented to enhance the established knowledge system. Therefore, whether appropriate knowledge is transferred to leverage related learning is an important evaluation metric related to CL.

According to the level of transferred information, the memory transfer techniques can be mainly divided into two categories. The first category involves directly transferring the outputs of old tasks, and the second category involves transferring hidden information such as feature maps, similarities and attributes across tasks.

\begin{table*}[!t]
\renewcommand{\arraystretch}{1.4}
\caption{Summary on information transfer method.}
\label{table_4}
\centering
\begin{tabular}{|c|c|c|c|}
\hline
Method & Algorithm & Keypoint & Reference\\
\hline
\multirow{2}{*}{Direct result transfer} & LWF & Knowledge distillation loss & [95]\\
 & iCaRL  & Reproduce old results & [36]\\
\hline
\multirow{5}{*}{Relevant experience transfer} & LFL & Transfer hidden activations  & [32]\\
& P\&C & Knowledge base  & [96] \\
& PNN & Reuse low-level visual features  & [55] \\
&  Expert Gate  &  Gating autoencoders & [97] \\
& A-GEM & Task descriptors & [98]\\
\hline
\end{tabular}
\end{table*}

The simplest way is to transfer the final results of previous tasks. For example, learning without forgetting (LwF) \cite{li2017learning} records the output of old tasks for new data and applies a knowledge distillation loss 0 as a regularizer to minimize the bias of stored outputs for old tasks while adapting to the new task. iCaRL \cite{rebuffi2017icarl} forces the model to reproduce the results of old classes when the model is updated for the new task, and it uses an exemplar set to store representative exemplars of old tasks. Similarly, EEIL \cite{castro2018end}, LSIL \cite{wu2019large}, and IL2M \cite{belouadah2019il2m} transfer the previous information in classification layers for the old classes.

However, retaining the results of past tasks in an invariant manner involves two problems: the model is prone to overfitting against the transferred examples, and the low relatedness among the tasks significantly increases the amount of forgetting. To exploit the useful information, certain methods propose transferring the relevant experience among tasks. Several algorithms transfer the previous feature information; for example, LFL \cite{jung2016less} regularizes the $l_2$ distance between the final hidden activations and preserves the previously learned input-output mappings by computing additional activations with the parameters of the old tasks. In this way, the target network learns to extract features that are similar to the features extracted by the source network. Progress \& compress (P\&C) \cite{schwarz2018progress} introduces a knowledge base to transfer the learned features. The learned knowledge is constantly distilled into the knowledge base in the compression phase, and the learned features are selected and reused for new learning in the progress phase. PNN \cite{rusu2016progressive} supports the transfer of features across sequences of tasks. This approach enables the reuse of all aspects from low-level visual features to high-level policies to facilitate transfer to new tasks by utilizing lateral connections to previously learned models. In this manner, PNN achieves a richer composition, in which prior knowledge is no longer transient and can be integrated in each layer of the feature hierarchy. Expert Gate \cite{aljundi2017expert} selects the most appropriate specialist model when learning new tasks. This approach introduces gating autoencoders that inherently decide which of the most relevant prior models is to be used for training a new expert and captures the relatedness between tasks. A dubbed averaged GEM (A-GEM) \cite{chaudhry2018efficient} introduces task descriptors that describe the attributes of a task. The task descriptors are shared across tasks, and thus, a model can promptly recognize a new class provided a descriptor specifies certain attributes that it has learned.

Information transfer is a useful technique in CL, and it can be flexibly embedded into many CL algorithms to enhance the efficiency of the learning process. Current memory transfer techniques in the memory retrospection method migrate the previous information pertaining to the updated model to consolidate past memory, facilitating BWT and FWT. The summary on these transfer techniques is presented in Table \ref{table_4}.

\section{Application}
A reliable and robust application is expected to pass tests in real-world scenarios in which data are dynamic and variable. Traditional training paradigms for models are confined in a fixed and static setting, resulting in their poor reaction to new surroundings. Therefore, to realize model applications, it is necessary to address the conflict between the inadequate response and demand for flexible adjustment according to different environments. The enhancement of the reaction is based on the preservation and utilization of learned experiences and efficient acquisition of new knowledge. In this context, CL is a promising and appropriate learning paradigm for models in applications. This chapter provides an overview of the major applications in which CL plays an important role.

\subsection{Object detection}
With the increasing deployment of object detection globally, the variety and novelty of objects to be detected are expected to increase beyond the designated range of classes, requiring the existing models to be updated to detect additional classes. Incremental object detection can identify and localize additional instances of semantic objects of a certain class. This approach broadly consists of selecting region proposals, extracting features, and predicting the object class for new objects while preserving the old objects.

The early work in this domain focused on the continuous location of region proposals. For example, Shmelkov et al. \cite{shmelkov2017incremental} used knowledge distillation to measure the discrepancy of distillation proposals between the old and new networks. This discrepancy is added as an additional term in the standard cross-entropy loss function for new classes. By minimizing the final loss function, the model can balance the interplay between the detection of old and new classes. CIFRCN \cite{hao2019end} involves an end-to-end class incremental detection method based on distilling of the RPN \cite{ren2015faster}. Continuous bounding box identification through RPN accelerates incremental object detection. Chen et al. \cite{chen2019new} note that the confidence in the coarse labels of the initial model contains abundant knowledge of the learned class and can help the model retain the old class information in the ROI. Therefore, the authors innovatively use the confidence loss to extract the confidence information of the initial model to suppress forgetting.

Certain approaches extensively consider continual feature extraction. For example, RILOD \cite{li2019rilod} introduces an extra feature distillation loss in addition to the classification and bounding box loss. Feature distillation prevents dramatic changes in the features extracted from a middle neural network layer, which further reduces catastrophic forgetting and increases the model accuracy. RODEO \cite{acharya2020rodeo} involves migrating the replay strategy to overcome forgetting in incremental object detection. This approach uses a memory buffer to store compressed representations of feature maps of old examples. Next, the new input is combined with a reconstructed subset of samples from the replay buffer when training the model over the new examples.

Certain approaches emphasize the continual classification loss in object detection because it is considered that the classifier misclassifies objects of old classes that lack class annotations as background when the model is adapted to new class images. IncDet \cite{liu2020incdet} designs the pseudobounding box annotation of previously learned classes to replace the class annotations. Specifically, for each image in new datasets, InceDet predicts old class objects with the old model and obtains pseudobounding boxes for old classes. Next, the old class pseudobounding boxes and new class annotated bounding boxes are jointly used to incrementally finetune the model.

Overall, object detection fully exploits the existing CL methods in combination with knowledge distillation techniques. These incremental object detection algorithms have achieved promising results, in the context of single-stage networks such as RetinaNet \cite{lin2017focal} and YOLO \cite{redmon2016you}\cite{redmon2017yolo9000} and two-stage networks such as Fast RCNN \cite{wang2017fast} and Faster RCNN \cite{ren2015faster}. Most of the existing studies focus on maintaining satisfactory information retrospection and information transfer ability; however, with the deepening of research, the memory prospection problem has been gradually recognized. For example, IncDet \cite{liu2020incdet} tests the mAP for new classes by adding them simultaneously to test the intransigence and LCA ability of the model.

\subsection{Image segmentation}
Incremental image segmentation is aimed at realizing pixel-level labeling with a set of continuous object categories \cite{michieli2019incremental}. In contrast to object detection, in continuous semantic segmentation, each image contains pixels belonging to multiple classes, and thus, the labeling is dense. In addition, the pixels associated with the background during a learning step may be assigned to a specific object class in subsequent steps or vice versa, making the problem highly complicated. Therefore, an effective image segmentation model must classify objects in a fine-grained manner and handle background classes sequentially in addition to the standard requirement for incremental object detection. Most incremental objection detection approaches realize memory retrospection by exploiting the transfer learning technique.

Michieli et al. \cite{michieli2019incremental} use a masked cross-entropy loss between the logits produced by the output layer in the previous model and current model to transfer the final results and apply $l_2$ loss to transfer the intermediate level of the feature space. This work is similar to the mainstream methods associated with incremental object detection.

Certain approaches divide traditional transfer learning into more detailed subprocesses. Mazen et al. \cite{mel2020incremental} propose a coarse-to-fine semantic segmentation method. This approach transfers the previously gained knowledge, acquired on a simple semantic segmentation task with coarse classes, to a new model involving more fine-grained and detailed semantic classes. This approach exploits domain sharing at the feature extraction level and can thus provide insight for seeking commonalities and differences across tasks in incremental segmentation tasks.

Notably, the existing studies have not considered the incremental transfer of background information. Cermelli et al. \cite{cermelli2020modeling} report that the semantics associated with the background class change over time and may be assigned to a specific object class in subsequent learning, causing the inaccuracy of segmentation results. To address this problem, the researchers enhance the traditional cross-entropy and distillation loss by considering the background information and initialized the classifier parameters to prevent biased predictions toward the background class.

Another problem in transfer learning is the imbalance problem since classes in the training data are highly imbalanced in most cases when determining the training patches to be transferred. Tasar et al. \cite{tasar2019incremental} alleviate the sample imbalance problem by selecting previous training data with high importance values. This method can achieve excellent segmentation results for remote sensing data.

Klingner et al. \cite{klingner2020class} note that the existing approaches are either restricted to settings in which the additional classes have no overlap with the old ones or rely on labels for both old and new classes. The authors introduce a generally applicable technique that learns new data solely from labels for the new classes and outputs of a pretrained teacher model. The segmentation model is trained in stage 1 on dataset 1 for a set 1 of classes. Later, the model is extended in stage 2 by additional classes 2 and dataset 2, such that the model outputs both old and new classes.

Generally, the process of incremental image segmentation is similar to that of incremental objection detection, with a higher emphasis on updating background classes. Currently, transfer learning is the most widely applied method in the specific field, and it can achieve promising results in information retrospection.

\subsection{Face recognition}
Face recognition \cite{jain2011handbook} is widely implemented in real-world situations, such as e-passports, ID cards and entrance guarding frameworks. Owing to the increasing number of people involved and facial changes due to aging, sunlight, angle, occlusion, expressions, and make-up, the original or static face database is either insufficient or inappropriate for future events. A solution to this problem is to ensure that face recognition systems learn continuously to accommodate unknown faces and known faces with spatial and temporal variation. A substantial number of algorithms for continuous face detection were introduced twenty years ago, and many of them achieved promising results in memory retrospection, memory prospection and memory transfer.

To facilitate memory retrospection, Toh et al. \cite{toh2003face} present an adaptive face recognition system based on memory storage. The initial idea is to accumulate and retrain all previous data when training new face images; however, this aspect is infeasible for large datasets. Therefore, the approach focuses only on misclassified samples. The approach uses RAN-LTM \cite{platt1991resource} to store the weights in hidden layers and create a memory item for misclassified samples. These pairs are trained with newly given training data to suppress forgetting. Kim et al. \cite{kim2012incremental} propose an incremental method for classic short-term memory (STM) and long-term memory (LTM) algorithms for face recognition. In this approach, the STM uses a recall algorithm to recall existing data from the LTM before incrementing new incoming data. La Torre et al. \cite{de2012incremental} report an incremental adaptive ensemble of classifiers in which new samples are trained and combined with the previously trained classifiers by using an iterative boolean combination (IBC).

Many researchers have focused on memory prospection for the latter task. For example, Chen et al. \cite{chen2008novel} adopt a constraint block NMF (CBNMF) to rapidly learn additional samples as well as additional class labels by increasing the between-class distances while reducing within-class distances based on a divergence criterion. Jia et al. \cite{jia2009incremental} propose an incremental version of a manifold learning technique known as Laplacian eigenmaps, in which the adjacency matrix is updated, followed by the projection of the new sample onto the lower dimensional space. This fast and efficient feature extraction technique accelerates the subsequent learning. Ye and Yang \cite{ye2015incremental} propose an incremental sparse representation classification (SRC) strategy. When new data arrive, the groups or classes to which these batch data belong are updated, while the remaining classes are retained. The high efficiency of this approach in updating the local dictionaries of the related class during training accelerates the learning process. Zhao et al. \cite{zhao2006novel} state that the existing incremental principal component analysis (IPCA) \cite{jeong2009ipca} methods cannot manage the approximation error, which deteriorates the results of the subsequent recognition. To address this limitation, the authors proposed SVDU-IPCA based on the idea of an SVD \cite{zha1999updating} updating algorithm. SVDU-IPCA achieves excellent face recognition results with an extremely small degradation in the accuracy.

To flexibly transfer the knowledge learned in the training process, Hisada et al. \cite{hisada2010incremental} extend the incremental linear discriminant analysis (LDA) \cite{balakrishnama1998linear} for the multitask recognition problem. Several classifications are performed with the same input; the knowledge from each task is extended to the remaining tasks and incrementally updated over the incoming inputs. The proposed algorithm can perform several levels of recognition, such as name, age, and gender. Sakai \cite{sakai2008monte} propose an incremental subspace generation approach known as the Monte Carlo subspace method that uses a row-incremental singular value decomposition (RiSVD) to select useful a priori knowledge to support current learning. Liwicki et al. \cite{liwicki2012efficient} propose an incremental PCA over the Krein space \cite{langer2004krein}, which can provide valuable insights for the classifier geometry. In the incremental update, the existing Krein subspace is used to create a complementary subspace based on the mapping functions of the new data. The correlation between the old and new inputs significantly enhances the recognition performance. Kang and Choi \cite{kang2006kernel} suggest an incremental version of the support vector domain description (SVDD) \cite{tax1999support} to find a minimum volume sphere that could enclose all or most of the data. In the incremental version, the authors proposed utilizing new data that contribute to the existing class descriptions to update the support vectors and discarding the remaining data.

In conclusion, incremental face recognition has been fully exploited. First, several techniques have been proposed to enhance memory retrospection, memory prospection and memory transfer. Second, in addition to traditional CL methods, novel strategies from other domains have been integrated. Third, many detailed and specific problems have been solved, for example, targeted to illuminance invariance. Furthermore, the reduction in the computational complexity of incrementation has been considered \cite{guan2012online}\cite{sakai2008monte}, which is of significance for large face datasets.

\subsection{Image generation}
Image generation refers to the task of generating new images from an existing dataset by learning high-dimensional data distributions. Image generation is widely implemented in fields with low data availability. With the requirement of diversified and big data, a model that can consecutively generate data according to different scene requirements is desirable. Continual image generation aims to continually generate new categories of images by remembering the old generation process \cite{gregor2015draw}. A robust continuous image generation model can capture the difference between multiple datasets in terms of the data distribution. Many strategies to protect memory retrospection, such as regularization and generative replay methods, have been applied.

\begin{figure*}
\centering
\includegraphics[width=12cm]{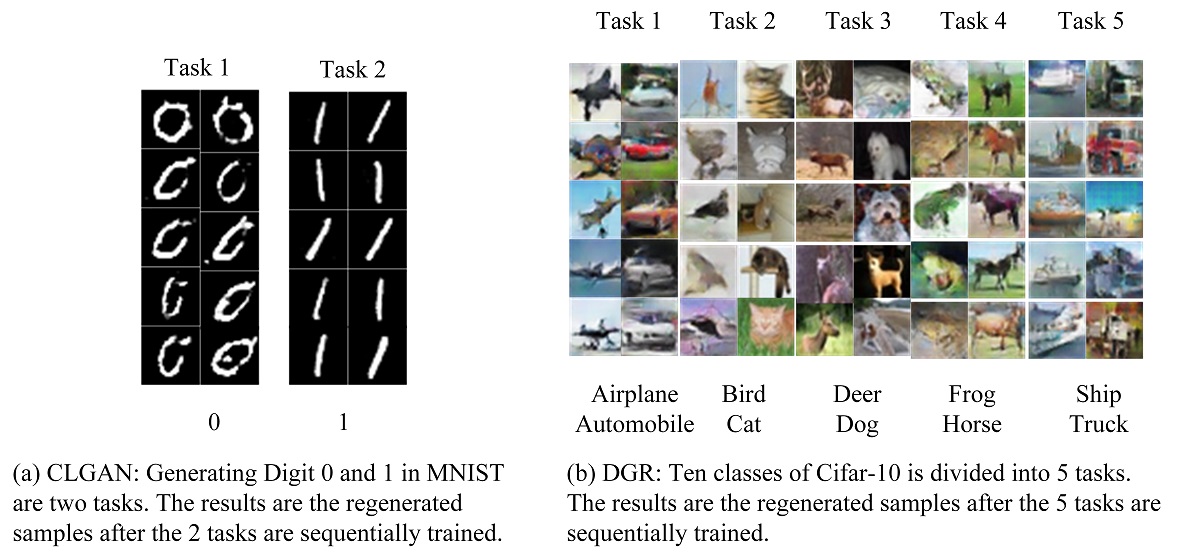}
\caption{Incremental image generation results of CLGAN with MNIST and DGR with Cifar-10.}
\label{img8}
\end{figure*}

In terms of regularization methods, CLGAN \cite{seff2017continual} is a pioneering and representative approach. This technique employs EWC in the generator of a GAN to assess the weight importance and elastically updates the weights in the generator when learning new information. Weights are confined to a region benefiting the performance for all data classes to prevent forgetting. However, it is difficult for CLGAN to seek a shared feature space when the data distribution varies significantly or the number of tasks is large. The performance of CLGAN is tested and shown in Figure \ref{img8} (a).

The generative replay method is commonly used in image generation because the generator can reproduce past pseudosamples. DGR \cite{shin2017continual} regenerates data and their corresponding responses on the past model as a pair to represent old tasks, and the pairs are interleaved with new data to update the model. The performance of CLGAN is tested and shown in Figure 8 (b). Certain modern approaches focus on leveraging the quality of regenerated images. For example, MeRGAN \cite{wu2018memory} uses CGAN, in which the category is used as an input to guide the replay process, thereby avoiding less biased sampling on past categories and generating more reliable past data. CloGAN \cite{rios2018closed} uses an auxiliary classifier to filter a portion of distorted replay to block inferior images from entering the training loop. BIR \cite{van2020brain} replays the internal or hidden representations of past tasks rather than the data. The replayed representation is generated by the context modulated feedback connection of the network. Similar to the replay idea, certain approaches reuse compressed information to represent old tasks rather than repetitive old data. DGM \cite{ostapenko2019learning} introduces task-specific binary masks to the generator weights when training the model on a class of images. These approaches preserve the mask instead of the data as the old clues. Two variants of DGM, known as DGMw and DGMa \cite{ostapenko2019learning}, have been proposed to ensure sufficient model capacity to accommodate incoming tasks by an adaptive network expansion mechanism.

In addition to memory retrospection, memory transfer has been considered. Memory GAN \cite{cong2020gan} creates mFiLM and mAdaFM modules to transfer the source information (old task) in the fully connected and convolutional layers, respectively, to the target domains (new task). The mFiLM and mAdaFM modules are designed for each individual task, and their parameters can be compressed to save memory, thereby allowing a long sequence of incremental tasks to be realized. Similarly, Piggyback GAN \cite{zhai2020piggyback} adopts piggyback filters in which a set of convolutional and deconvolutional filters of previous tasks are factorized. Piggyback GAN can be scale up with the tasks since the filters for previous tasks are not altered and can be restored in a piggyback bank. In addition, the approach leverages weight reuse and adaptation to increase the parameter efficiency when extending GAN to new tasks. GRFC \cite{van2018generative} also utilizes a distillation strategy to store past information and reduces the computational cost by integrating the generative model in the main model through generative feedback connections.

More complex tasks such as image-to-image translation have been explored. LiSS \cite{schmidt2020towards} achieves continual unpaired image-to-image translation on CycleGAN \cite{zhu2017unpaired}. The approach distills the knowledge of a reference encoder, which is an exponential moving average of previous encoders in the parameter space, to assist CycleGAN in more effectively disentangling the instances of the objects to be translated. Since the reference encoder can maintain a weighted memory of all the past encoders at the cost of a single additional encoder, LiSS requires low memory and computational cost.

In general, many classic CL algorithms have been successfully applied to continual image generation, and the idea of a replay strategy has been widely studied. However, most of the existing research emphasizes the retention of the retrograde memory and neglects the anterograde memory. In addition, the existing techniques lack backward and forward knowledge transfer to boost bidirectional learning.

\subsection{Image fusion}
Image fusion encloses all data analysis strategies aiming at combining the information of several images obtained with the same platform or by different spectroscopic platforms \cite{cocchi2019data}. Since the source of images is always multiple, fusing different types of images is required, for example, magnetic resonance imaging (MRI), computerized tomography (CT), positron emission tomography (PET) and single-photon emission computed tomography (SPECT) modalities are always combined together to detect disease in medical operation \cite{du2016overview}; multi-spectral images and panchromatic images are always synthetized to detect abundant ground information in the remote sensing community \cite{wang2005comparative}.

Continual image fusion aims to solve different fusion problems, such as multi-modal, multi-exposure, multi-focus cases. Many DL methods have been putting forward to solve image fusion problem, but the continual image fusion work is far from satisfactory.

\begin{figure*}
\centering
\includegraphics[width=12cm]{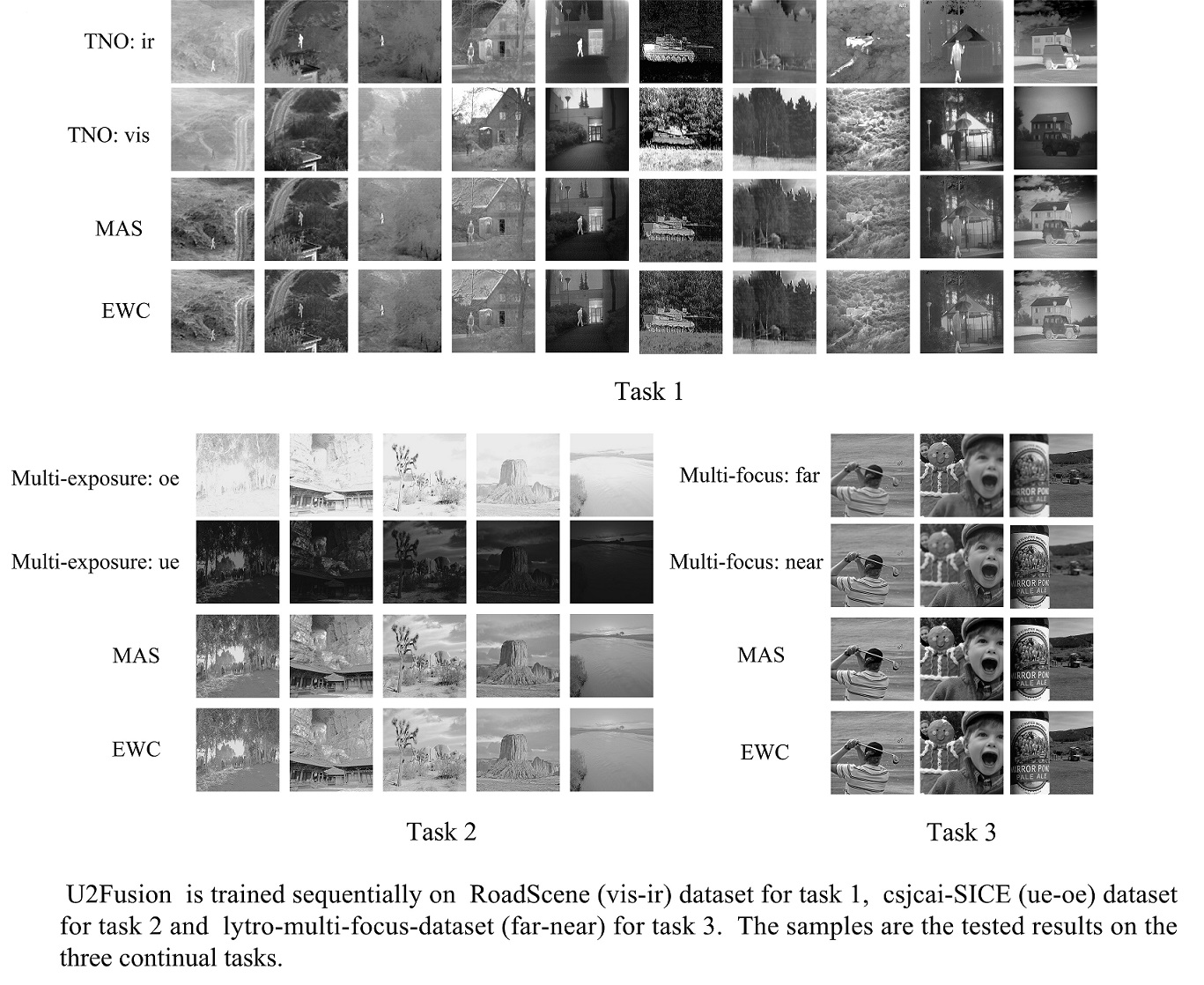}
\caption{Incremental image fusion of U2Fusion with three continuous tasks.}
\label{img9}
\end{figure*}

The latest and the only work is from U2Fusion \cite{xu2020u2fusion}, a unified model that is applicable to multiple fusion tasks. In U2Fusion, a feature extractor is first adopted to extract abundant and comprehensive features from source images. Then, the richness of information in features is measured to define the relative importance of these features by EWC. The relative importance represents the similarity relationship between the source images and the fusion result. A higher similarity entails that more information in this source image is preserved in the result, thus leading to a higher information preservation degree. We conducted the U2Fusion by its original EWC strategy and also tried the classical MAS method. Experiment results present that U2Fusion can achieve continual image fusion on multi-modal, multi-exposure, multi-focus cases. The performance of U2Fusion is tested as shown in Figure \ref{img9}.

\begin{figure*}
\centering
\includegraphics[width=12cm]{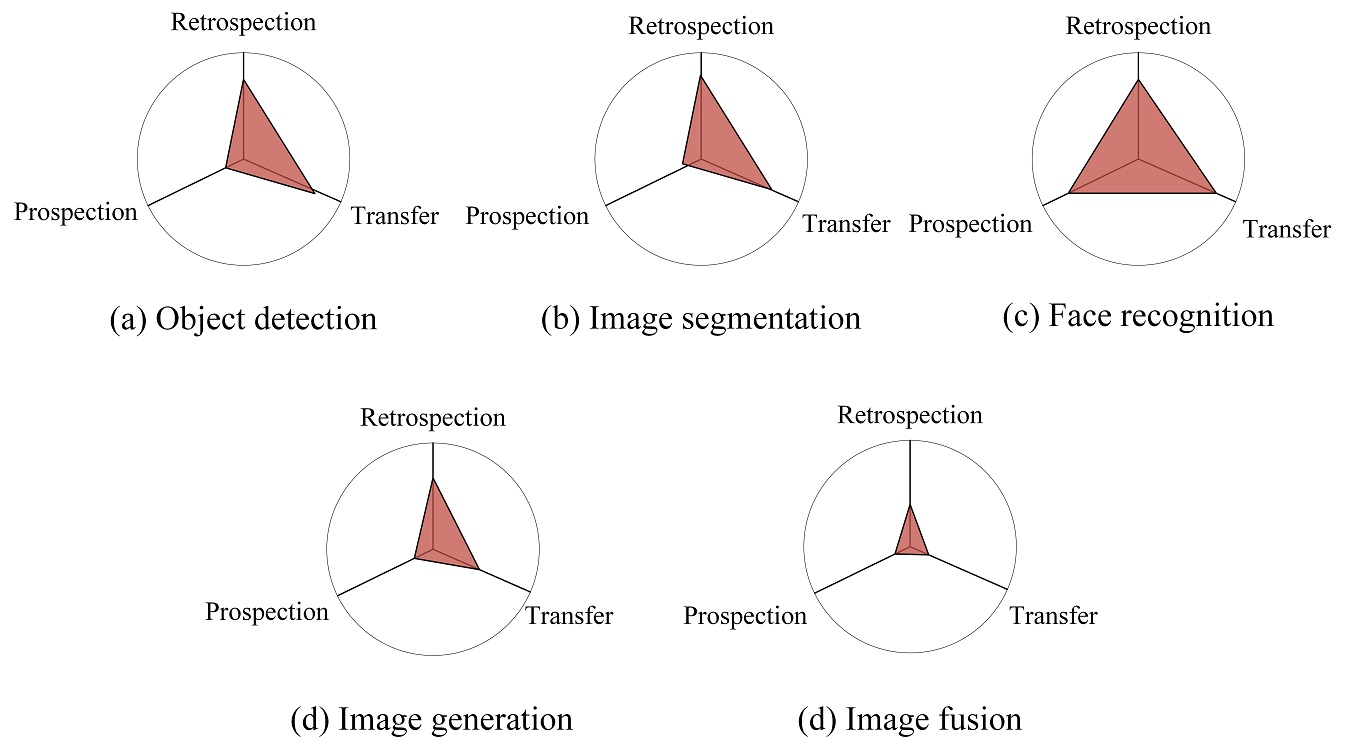}
\caption{Valuations of the major CL applications on the performance of information retrospection, information prospection and information transfer..}
\label{img10}
\end{figure*}

In conclusion, several applications were examined long ago, whereas others have been gradually considered in recent years. Therefore, the development level of these applications differs, and their performance on the information retrospection, information prospection and information transfer is shown in Figure \ref{img10}. Generally, with the combination of CL, the models are exhibiting a higher generalization ability in the open world and are thus more reliable and efficient in the relevant applications.

\section{Discussion}
Overall, CL has been fully developed and facilitated developments in other aspects of artificial intelligence. Considering the continual research advancements, we present several new concepts based on neurobiology knowledge.

To enhance memory retrospection, the following ideas may be considered.
\begin{itemize} \item \emph{Considering the chain reaction associated with the change in weights}. Enhancing the values of certain weights and dampening the action of neighboring weights in a model may help form a ¡°receptive field¡± for inputs, which can avoid a specific task from being interfered with. This insight is inspired by a neuroscience phenomenon termed "lateral inhibition" \cite{amari1977dynamics} that describes how an excited neuron reduces the activity of its neighbors. Lateral inhibition disables the spreading of action potentials from excited neurons to neighboring neurons in the lateral direction, which can create a contrast in the stimulation that allows an enhanced sensory perception. The increased stimulation can help the human brain form the initial memory. The existing CL methods individually and separately regard and adjust the activities of each weight. It must be examined whether the chain reaction resulting from the change in weights can improve memory retention.\\
\item \emph{The modality of past information should be innovated}. Most of the methods to protect retrograde memory use true or regenerated samples when referring to old data. However, the information that is processed in the brain is abstracted and compressed in the hippocampus \cite{burgess1998representation}\cite{pavlides1989influences}, which indicates that the modality of learned information is not as simple as its original input. In addition, condensed information can help models focus on essential information and save computational overhead. Therefore, enhancing the form of old information is a promising research direction.\\
\item \emph{The relevance of different past information should be considered}. When constructing a memory buffer for past memory, the existing methods mix the data. However, different categories of data may share similar low-level information and are mutually correlated in other aspects. In addition, biological research indicates that learned information is reorganized between the hippocampus and neocortex, with several neurons being weakened, strengthened and newly formed \cite{diba2007forward}. Therefore, the old information in the memory buffer should not be a simple integration of the original data and can be structured as a knowledge graph or in other modalities.\\
\item \emph{Preserving past information in an online way}. The existing approaches select or generate old samples after learning the task and before learning a new task, which is an offline technique. Recent neuroscience research shows that hippocampal replay not only exists in rest or sleep but also occurs during the learning of the current task \cite{rasch2013sleep}. Therefore, future work can examine techniques to leverage the process of retrograde memory preservation in an online and dynamic way, which can not only accelerate the training of the model but mimic human intelligence.
\end{itemize}

To facilitate memory prospection, the following innovative idea can be considered.
\begin{itemize} \item \emph{Considering additional "forgetting" to provide space for information prospection}. LTD, which represents forgetting, is as important as LTP, which represents the memorization of the synaptic plasticity, because our brain may be exhausted with the influx of information if no forgetting occurs. Moreover, the idea that forgetting might be beneficial for memory maintenance has been frequently expressed \cite{kingma2013auto}. Existing CL models always focus on memorizing important old information, and it must be examined whether active forgetting plays an important role in anterograde memory. By forgetting certain redundant information, the CL models may have additional room to learn new information and more flexibility to optimize this information.
\end{itemize}

To enable memory transfer, the following changes can be introduced in the model architecture:
\begin{itemize} \item \emph{Designing submodels based on the main CL models to regulate information transfer}. The CLS theory states that the hippocampal system exhibits short-term adaptation and enables the rapid learning of novel information that is played back over time to the neocortical system for long-term retention \cite{mcclelland1995there}. The interplay of hippocampal and neocortical functionalities is crucial to transfer different memories. The cooperation of different areas in the brain indicates that the expansion of the structure of the model is not necessarily restricted to the original one. In fact, more concurrent submodels can be designed to regulate memory circulation in combination with the main models.
\end{itemize}

\section{Conclusion}
CL has attracted considerable attention in the deep learning community, and many effective algorithms have been proposed to achieve CL in ANNs. We reconsider the learning of CL in three aspects: memory retrospection, memory prospection and memory transfer, homologous to the human CL system. Consolidating information retrospection is a prerequisite in CL; information prospection represents the fast acquisition and adaptation of novel knowledge; flexibly transferring information promotes the efficiency and intelligence of a CL agent. Although significant progress and achievements have been witnessed in CL, additional illuminating ideas deserve to be introduced. Rethinking CL from the perspective of neuroscience knowledge can help design CL agents that can mimic human intelligence.

\bibliography{bare_jrnl_new_sample4}

\end{document}